\documentclass{article} % For LaTeX2e
\usepackage{nips12submit_e,times}
\usepackage{multirow}
\usepackage{amsfonts}
\usepackage{amssymb}
\usepackage{amsmath}
\usepackage{subfigure}
\usepackage{graphicx}
\usepackage{mathrsfs}
\usepackage{algorithm}
\usepackage{algorithmic}

\def\1{\boldsymbol{1}}

\title{A Nested HDP for Hierarchical Topic Models}

\author{
John Paisley\\
Department of EECS\\
UC Berkeley \\
\And
Chong Wang \\
Dept. of Machine Learning \\
Carnegie Mellon University\\
\And
David Blei\\
Dept. of Computer Science\\
Princeton University\\
\And
Michael I. Jordan\\
Department of EECS\\
UC Berkeley\\
}

\nipsfinalcopy % Uncomment for camera-ready version

\begin{document}

\maketitle

\begin{abstract}
We develop a nested hierarchical Dirichlet process (nHDP) for hierarchical topic modeling. The nHDP is a generalization of the nested Chinese restaurant process (nCRP) that allows each word to follow its own path to a topic node according to a document-specific distribution on a shared tree. This alleviates the rigid, single-path formulation of the nCRP, allowing a document to more easily express thematic borrowings as a random effect. We demonstrate our algorithm on 1.8 million documents from \emph{The New York Times}.\footnote{This is a workshop version of a longer paper. Please see \cite{Paisley:2013} for more details, including a scalable inference algorithm.}
\end{abstract}

\section{Introduction}
Organizing things hierarchically is a natural process of human activity. A hierarchical tree-structured representation of data can provide an illuminating means for understanding and reasoning about the information it contains. The nested Chinese restaurant process (nCRP) \cite{Blei:2010} is a model that performs this task for the problem of topic modeling.  It does this by learning a tree structure for the underlying topics, with the inferential goal being that topics closer to the root are more general, and gradually become more specific in thematic content when following a path down the tree.

\begin{figure}[t]\centering
 \includegraphics[width=.7\textwidth]{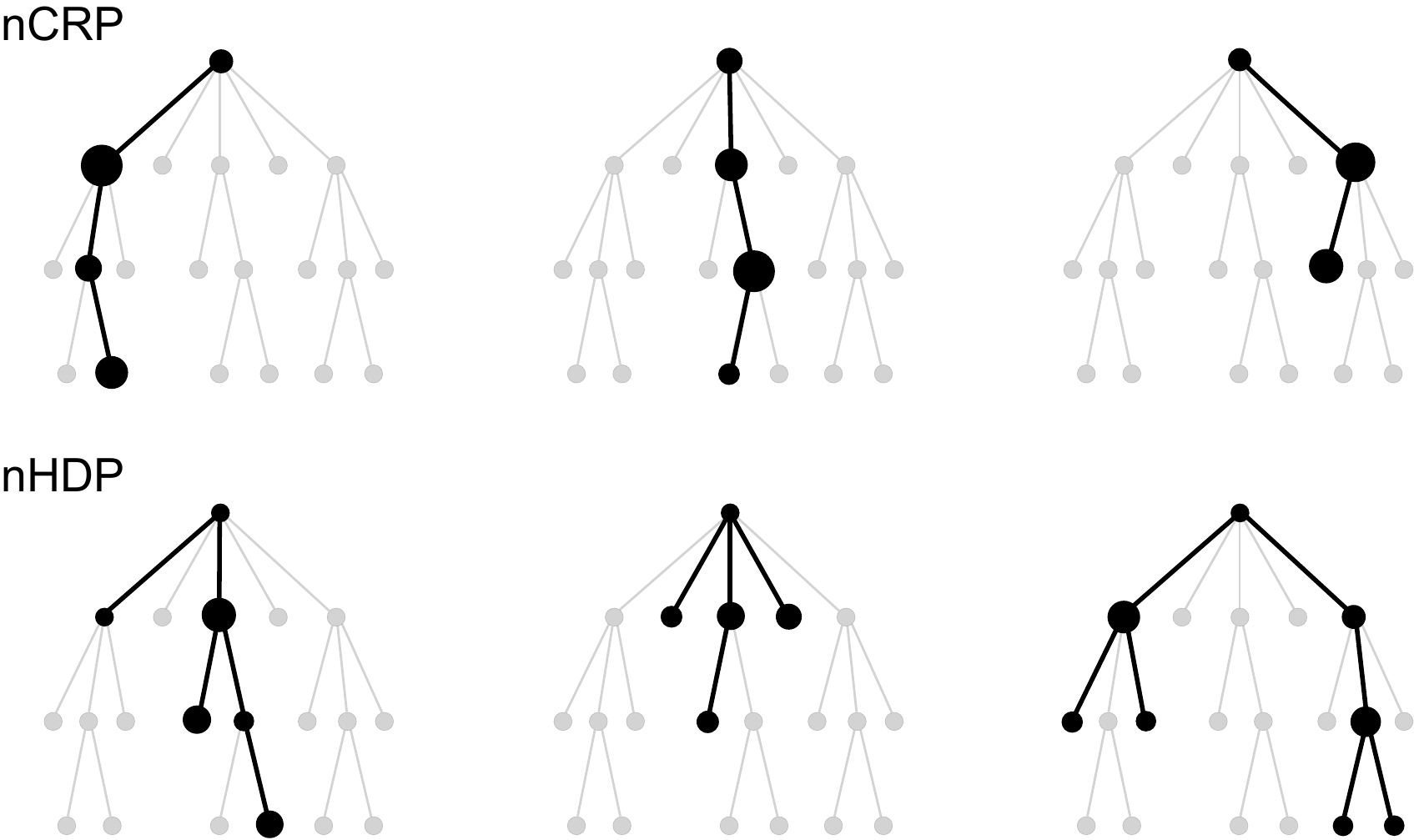}
\caption{An example of path structures for the nested Chinese restaurant process (nCRP) and the nested hierarchical Dirichlet process (nHDP) for hierarchical topic modeling. With the nCRP, the topics for a document are restricted to lying along a single path to a root node. With the nHDP, each document has access to the entire tree, but a document-specific distribution on paths will place high probability on a particular subtree. The goal of the nHDP is to learn a thematically consistent tree as achieved by the nCRP, while allowing for the cross-thematic borrowings that naturally occur within a document. \label{fig.toy_trees}}
\end{figure}

The nCRP is limited in the hierarchies it can model. We illustrate this limitation in Figure \ref{fig.toy_trees}. The nCRP models the topics that go into constructing a document as lying along one path of the tree. This significantly limits the underlying structure that can be learned from the data. The nCRP performs \emph{document-specific} path clustering; our goal is to develop a related Bayesian nonparametric prior that performs \emph{word-specific} path clustering. We illustrate this objective in Figure \ref{fig.toy_trees}. To this end, we make use of the hierarchical Dirichlet process \cite{Teh:2006}, developing a novel prior that we refer to as the \emph{nested hierarchical Dirichlet process} (nHDP). With the nHDP, a top-level nCRP becomes a base distribution for a collection of second-level nCRPs, one for each document. The nested HDP provides the opportunity for cross-thematic borrowing that is not possible with the nCRP. 

\section{Nested Hierarchical Dirichlet Processes}

\textbf{Nested CRPs.\,\,}
Nested Chinese restaurant processes (nCRP) are a tree-structured extension of the CRP that are useful for hierarchical topic modeling. A natural interpretation of the nCRP is as a tree where each parent has an infinite number of children. Starting from the root node, a path is traversed down the tree. Given the current node, a child node is selected with probability proportional to the previous number of times it was selected among its siblings, or a new child is selected with probability proportional to a parameter $\alpha > 0$. 

For hierarchical topic modeling with the nCRP, each node has an associated discrete distribution on word indices drawn from a Dirichlet distribution. Each document selects a path down the tree according to a Markov process, which produces a sequence of topics. A stick-breaking process provides a distribution on the selected topics, after which words for a document are generated by first drawing a topic, and then drawing the word index.

\textbf{HDPs.\,\,}\label{sec.HDP}
The HDP is a multi-level version of the Dirichlet process. It makes use of the idea that the base distribution for the DP can be discrete. A discrete base distribution allows for multiple draws from the DP prior to place probability mass on the same subset of atoms. Hence different groups of data can share the same atoms, but place different probability distributions on them. The HDP draws its discrete base from a DP prior, and so the atoms are learned.

\textbf{Nested HDPs.\,\,}\label{sec.nHDP}
In building on the nCRP framework, our goal is to allow for each document to have access to the entire tree, while still learning document-specific distributions on topics that are thematically coherent. Ideally, each document will still exhibit a dominant path corresponding to its main themes, but with offshoots allowing for random effects. Our two major changes to the nCRP formulation toward this end are that ($i$) each word follows its own path to a topic, and ($ii$) each document has its own distribution on paths in a shared tree.

With the nHDP, all documents share a global nCRP. This nCRP is equivalently an infinite collection of Dirichlet processes with a transition rule between DPs starting from a root node. With the nested HDP, we use each Dirichlet process in the global nCRP as a base for a second-level DP drawn independently for each document. This constitutes an HDP. Using a nesting of HDPs allows each document to have its own tree where the transition probabilities are defined over the same subset of nodes, but where the values of these probabilities are document-specific. 

For each node in the tree, we also generate document-specific switching probabilities that are i.i.d.\ beta random variables. When walking down the tree for a particular word, we stop with the switching probability of the current node, or continue down the tree according to the probabilities of the child HDP. We summarize this process in Algorithm \ref{alg.nHDP}.

\begin{algorithm}[b]
   \caption{Generating Documents with the Nested Hierarchical Dirichlet Process}
   \label{alg.nHDP}
\raggedright\vspace{4pt}
\begin{description}
\item{Step 1.} Generate a global tree by constructing an nCRP.
\item{Step 2.} For each document, generate a document tree from the HDP and switching probabilities for each node.
\item{Step 3.} Generate a document. For word $n$ in document $d$,\\
   a) Walking down the tree using HDPs. At current node, continue/stop according to its switching probability.\\
   b) Sample word $n$ from the discrete distribution at the terminal node.
\end{description}
\end{algorithm}

\begin{figure}[h!]\centering \vspace{-.5cm}
 \includegraphics[width=1\textwidth]{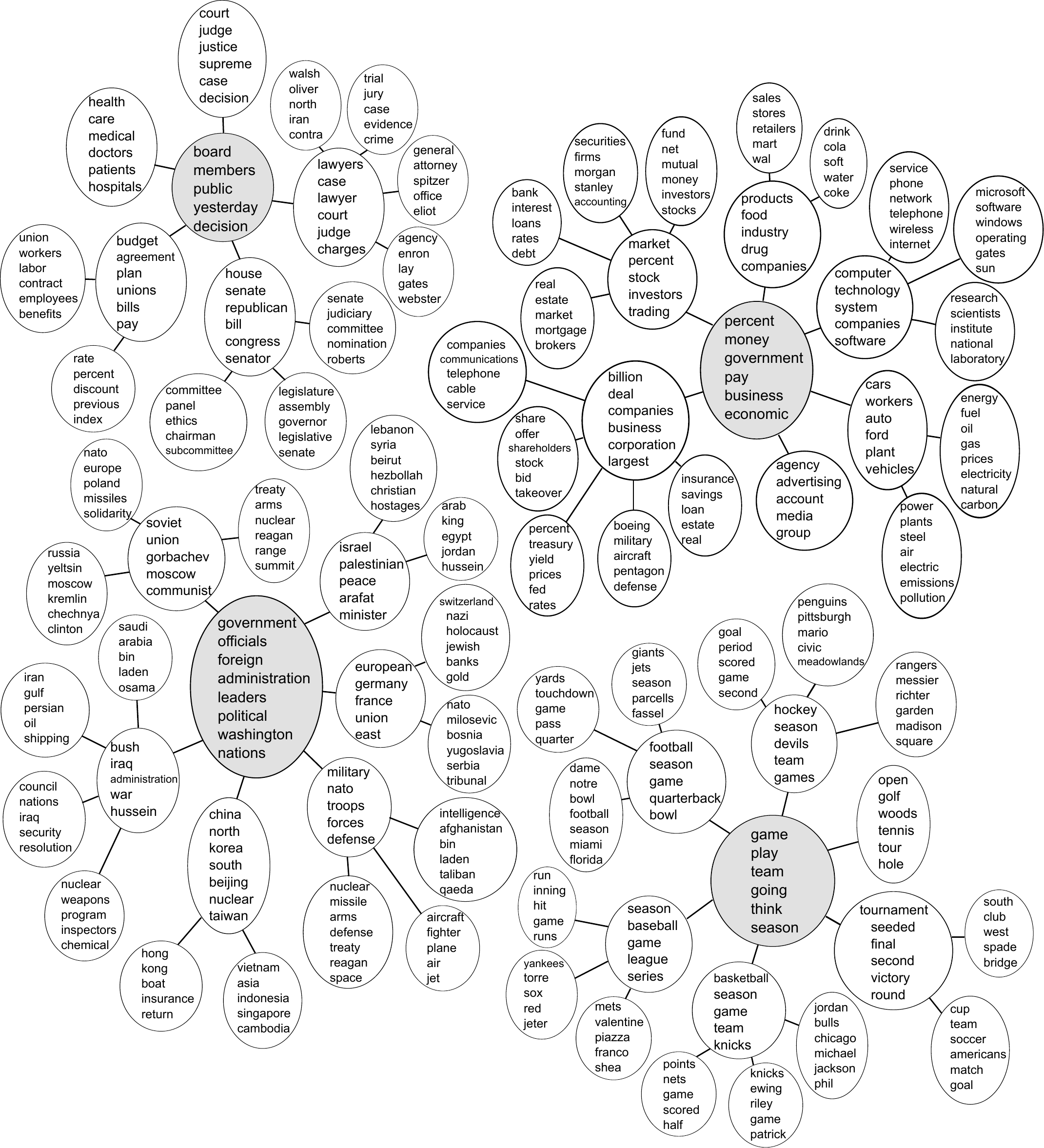}
\caption{Tree-structured topics from The New York Times. A shaded node is a top-level node.\label{fig.nyt_topics}}
\end{figure}

\textbf{Sample Results.\,\,}\label{sec.experiments}
We present some qualitative results for the nHDP topic model on a set of 1.8 million documents from \emph{The New York Times}. These results were obtained using a scalable variational inference algorithm \cite{Hoffman:2012} after one pass through the data set. In Figure \ref{fig.nyt_topics} we show example topics from the model and their relative structure. We show four topics from the top level of the tree (shaded), and connect topics according to parent/child relationship. The model learns a meaningful hierarchical structure; for example, the sports subtree branches into the various sports, which themselves appear to branch by teams. In the foreign affairs subtree, children tend to group by major subregion and then branch out into subregion or issue.

\bibliographystyle{IEEEtran}  
\bibliography{pami_bnp}

\end{document}